\documentclass[runningheads]{llncs}

\usepackage[T1]{fontenc}

\usepackage{cite}
\usepackage{url}
\usepackage[pdftex]{graphicx}

\usepackage{tikz}
\usepackage{pgfplots}
\pgfplotsset{compat=newest}
\usepgfplotslibrary{external}
\usetikzlibrary{pgfplots.groupplots,external}
\usepackage[cmex10]{amsmath}
\usepackage{amsfonts}

\usepackage{mathtools}
\usepackage{algorithm}
\usepackage{algpseudocode}
\usepackage{xspace}
\usepackage{soul}

\usepackage{graphicx}
\usepackage{mathrsfs}
\usepackage{mathtools}
\usepackage{changepage}

\usepackage{multirow}

\allowdisplaybreaks

\def\HiLi{\leavevmode\rlap{\hbox to .94\linewidth{\color{yellow!50}\leaders\hrule height .8\baselineskip depth .5ex\hfill}}}
\def\HiLii{\leavevmode\rlap{\hbox to .91\linewidth{\color{yellow!50}\leaders\hrule height .8\baselineskip depth .5ex\hfill}}}

\newcommand{\R}{\mathbb{R}}

\newcommand{\X}{\ensuremath{\mathrm{X}}\xspace} 
\newcommand{\Y}{\ensuremath{\mathrm{Y}}\xspace} 
\newcommand{\domain}{\ensuremath{S}\xspace} 
\newcommand{\E}{\ensuremath{\operatorname{E}}} 
\newcommand{\af}{\ensuremath{\mathcal{\alpha}}} 
\renewcommand{\H}{\ensuremath{\mathcal{H}}\xspace} 

\newcommand{\inner}[2]{\ensuremath{\left\langle #1, #2\right\rangle_\H}} 
\newcommand{\norm}[1]{\ensuremath{\left\lVert#1\right\rVert}} 
\newcommand{\Span}[1]{\ensuremath{\operatorname{span}\{#1\}}} 
\newcommand{\rank}{\ensuremath{\operatorname{rank}}}

\renewcommand{\epsilon}{\varepsilon}

\renewcommand{\Pr}{\mathbb{P}}

\DeclareMathOperator*{\argmin}{arg\,min}
\DeclareMathOperator*{\argmax}{arg\,max}

\newcommand{\BF}[1]{
	\relax
	\ifmmode
	\ifcat\noexpand#1\relax 
		\boldsymbol{#1}     
	\else
		\mathbf{#1}
	\fi
	\else
		\textbf{#1}
	\fi
}

\usepackage{fullpage}

\begin{document}

\setlength{\abovedisplayskip}{6pt}
\setlength{\belowdisplayskip}{6pt}

\title{High Dimensional Bayesian Optimization with Kernel Principal Component Analysis}
\titlerunning{High Dimensional Bayesian Optimization with Kernel PCA}

\author{Kirill Antonov\inst{1,3}\thanks{Work done while visiting Sorbonne Universit\'e in Paris. Corresponding author, \email{k.antonov@liacs.leidenuniv.nl}}\orcidID{0000-0002-8757-8598}
\and Elena Raponi\inst{2,4}\orcidID{0000-0001-6841-7409}
\and \\Hao Wang\inst{3}\orcidID{0000-0002-4933-5181} 
\and Carola Doerr\inst{4}\orcidID{0000-0002-4981-3227}
}

\institute{ITMO University, Saint Petersburg, Russia,
\and  Technical University of Munich, TUM School of Engineering and Design, Munich, Germany
\and Leiden University, LIACS Department, Leiden, Netherlands
\and Sorbonne Université, CNRS, LIP6, Paris, France}

\maketitle

\begin{abstract}
Bayesian Optimization (BO) is a surrogate-based global optimization strategy that relies on a Gaussian Process regression (GPR) model to approximate the objective function and an acquisition function to suggest candidate points. It is well-known that BO does not scale well for high-dimensional problems because the GPR model requires substantially more data points to achieve sufficient accuracy and acquisition optimization becomes computationally expensive in high dimensions. Several recent works aim at addressing these issues, e.g., methods that implement online variable selection or conduct the search on a lower-dimensional sub-manifold of the original search space. Advancing our previous work of PCA-BO that learns a linear sub-manifold, this paper proposes a novel kernel PCA-assisted BO (KPCA-BO) algorithm, which embeds a non-linear sub-manifold in the search space and performs BO on this sub-manifold. Intuitively, constructing the GPR model on a lower-dimensional sub-manifold helps improve the modeling accuracy without requiring much more data from the objective function. Also, our approach defines the acquisition function on the lower-dimensional sub-manifold, making the acquisition optimization more manageable.

We compare the performance of KPCA-BO to a vanilla BO and to PCA-BO on the multi-modal problems of the COCO/BBOB benchmark suite. Empirical results show that KPCA-BO outperforms BO in terms of convergence speed on most test problems, and this benefit becomes more significant when the dimensionality increases. For the 60D functions, KPCA-BO achieves better results than PCA-BO for many test cases. Compared to the vanilla BO, it efficiently reduces the CPU time required to train the GPR model and to optimize the acquisition function compared to the vanilla BO.
\end{abstract}


\sloppy{
\section{Introduction}\label{sec:intro}
Numerical black-box optimization problems are challenging to solve when the dimension of the problem's domain becomes high~\cite{bellman_dynamic_1966}. The well-known \textit{curse of dimensionality} implies that exponential growth of the data points is required to maintain a reasonable coverage of the search space. This is difficult to accommodate in numerical black-box optimization, which aims to seek a well-performing solution with a limited budget of function evaluations. Bayesian optimization (BO)~\cite{mockus1975bayesian,jones_efficient_1998} suffers from high dimensionality more seriously compared to other search methods, e.g., evolutionary algorithms, since it employs a surrogate model of the objective function internally, which scales poorly with respect to the dimensionality (see Sec.~\ref{sec:background} below). Also, BO proposes new candidate solutions by maximizing a so-called acquisition function (see Sec.~\ref{sec:background}), which assesses the potential of each search point for making progresses. The maximization task of the acquisition function is also hampered by high dimensionality. As such, BO is often taken only for small-scale problems (typically less than 20 search variables), and it remains an open challenge to scale it up for high-dimensional problems~\cite{binois_survey_2021}. 

Recently, various methods have been proposed for enabling high-dimensional BO, which can be categorized into three classes: (1) variable selection methods that only execute BO on a subset of search variables~\cite{UlmasovBODimScheduling}, (2) methods that leverage the surrogate model to high dimensional spaces, e.g., via additive models~\cite{duvenaud_additive_2011,delbridge_randomly_2020}, and (3) conducting BO on a sub-manifold embedded in the original search space~\cite{huang_scalable_2015,wang_bayesian_2016}. Notably, in~\cite{gaudrie_modeling_2020}, a kernel-based approach is developed for parametric shape optimization in computer-aided design systems, which is not a generic approach since the kernel function is based on the representation of the parametric shape and is strongly tied to applications in mechanical design. In~\cite{RaponiWBBD20} we proposed PCA-BO, in which we conduct BO on a linear sub-manifold of the search space that is learned from the linear principal components analysis (PCA) procedure.

This paper advances the PCA-BO algorithm by considering the kernel PCA procedure~\cite{scholkopf_nonlinear_1998}, which is able to construct a non-linear sub-manifold of the original search space. The proposed algorithm - \emph{Kernel PCA-assisted BO} (KPCA-BO) adaptively learns a nonlinear forward map from the original space to the lower-dimensional sub-manifold for reducing dimensionality and constructs a backward map that converts a candidate point found on the sub-manifold to the original search space for the function evaluation. We evaluate the empirical performance of KPCA-BO on the well-known BBOB problem set~\cite{hansen2020coco}, focusing on the multi-modal problems.

This paper is organized as follows. In Sec.~\ref{sec:background}, we will briefly recap Bayesian optimization and some recent works on alleviating the issue of high dimension for BO. In Sec.~\ref{sec:kernel-PCABO}, we describe the key components of KPCA-BO in detail. The experimental setting, results, and discussions are presented in Sec.~\ref{sec:experiement}, followed by the conclusion and future works in Sec.~\ref{sec:conclusion}.

\section{Related Work} \label{sec:background}
\paragraph{Bayesian Optimization (BO)}~\cite{jones_efficient_1998,ShahriariSWAF16} is a sequential model-based optimization algorithm which was originally proposed to solve single-objective black-box optimization problems that are expensive to evaluate. BO starts with sampling a small initial design of experiment (DoE, obtained with e.g., Latin Hypercube Sampling~\cite{daglib/0022270} or low-discrepancy sequences~\cite{niederreiter1988low}) $\X \subseteq \domain$. After evaluating $f(x)$ for all $x \in X$, it proceeds to construct a probabilistic model $\Pr(f \mid \X, \Y)$ (e.g., Gaussian process regression, please see the next paragraph). BO balances  exploration and exploitation of the search by considering, for a decision point $\BF{x}$, two quantities: the predicted function value $\hat{f}(\BF{x})$ and the uncertainty of this prediction (e.g., the mean squared error $\E(f(\BF{x}) - \hat{f}(\BF{x}))^2$). Both of them are taken to form the acquisition function $\af\colon \domain \rightarrow \R$ used in this work, i.e., the expected improvement~\cite{jones_efficient_1998}, which quantifies the potential of each point for making progresses. BO chooses the next point to evaluate by maximizing the acquisition function. After evaluating $\BF{x}^*$, we augment the data set with $(\BF{x}^*, f(\BF{x}^*))$ and proceed with the next iteration. \\
\paragraph{Gaussian Progress Regression} (GPR)~\cite{RasmussenW06} models the objective function $f$ as the realization of a Gaussian process $f \sim \operatorname{gp}(0, c(\cdot, \cdot))$, where $c\colon \domain\times\domain \rightarrow \R$ is the covariance function, also known as kernel. That is, $\forall\BF{x},\BF{x}'\in\domain$, it holds that $\operatorname{Cov}\{f(\BF{x}), f(\BF{x}')\} = c(\BF{x}, \BF{x}')$. Given a set $\X$ of evaluated points and the corresponding function values $\Y$, GPR learns a posterior Gaussian process to predict the function value at each point, i.e., $\forall \BF{x}\in\domain, f(\BF{x})\mid \X, \Y,\BF{x}\sim \mathcal{N}(\hat{f}(\BF{x}), \hat{s}^2(\BF{x}))$, where $\hat{f}$ and $\hat{s}^2$ are the posterior mean and variance functions, respectively.
When equipped with a GPR and the expected improvement, BO has a convergence rate of $O(n^{-1/d})$~\cite{Bull11}, which decreases quickly when the dimension increases.\\
\paragraph{High-dimensional Bayesian optimization.} 
High dimensionality negatively affects the performance of BO in two aspects, the quality of the GPR model and the efficiency of acquisition optimization. For the quality of the GPR model, it is well-known\cite{Bull11} that many more data points are needed to maintain the modeling accuracy in higher dimensions. Moreover, acquisition optimization is a high-dimensional task requiring more surrogate evaluations to obtain a reasonable optimum. Notably, each surrogate evaluation takes $O(d)$ time to compute, making the acquisition optimization more time-consuming.

Depending on the expected structure of the high-dimensional problem, various strategies for dealing with this curse of dimensionality have been proposed in the literature, often falling into one of the following classes:
\begin{enumerate}
    \item Variable selection or screening. It may be the case that a subset of parameters does not have any significant impact on solutions' quality, and it is convenient to identify and keep only the most influential ones. Different approaches may be considered: discarding variables uniformly \cite{LiBODropOut}, assigning weights to the variables based on the dependencies between them \cite{UlmasovBODimScheduling}, identifying the most descriptive variables based on their length-scale value in the model \cite{SalemSequentialDimReduction}, etc. 
    \item Additive models. They keep all the variables but limit their interaction as they are based on the idea of decomposing the problem into blocks. For example, the model kernels can be seen as the sum of univariate ones \cite{duvenaud_additive_2011, delbridge_randomly_2020}, the high-dimensional function can decompose as a sum of lower-dimensional functions on subsets of variables \cite{rolland_high-dimensional_2018}, or the additive model can be based on an ANOVA decomposition \cite{muehlenstaedt_data-driven_2012-1,ginsbourger_anova_2014}.
    \item Linear/nonlinear embeddings. They are based on the hypothesis that a large percentage of the variation of a high-dimensional function can be captured in a low-dimensional embedding of the original search space. The embedding can be either linear \cite{wang_bayesian_2016, RaponiWBBD20} or nonlinear \cite{guhaniyogi_compressed_2016, gaudrie_modeling_2020}. 
\end{enumerate}
We point the reader to~\cite{binois_survey_2021} for a comprehensive overview of the state-of-the-art in high-dimensional BO. 

\section{Kernel-PCA assisted by Bayesian Optimization}
\label{sec:kernel-PCABO}
In this paper, we deal with numerical black-box optimization problems $f\colon \domain\subseteq \mathbb{R}^d\rightarrow \mathbb{R}$, where the search domain is a hyperbox, i.e., $\domain=[l_1,u_1]\times[l_2,u_2]\times\cdots\times[l_d,u_d]$.
We reduce the dimensionality of the optimization problem \emph{on-the-fly}, using a \emph{Kernel Principal Component Analysis} (KPCA)~\cite{scholkopf_nonlinear_1998} for learning, from the evaluated search points, a non-linear sub-manifold $\mathcal{M}$ on which we optimize the objective function. Ideally, such a sub-manifold $\mathcal{M}$ should capture important information of $f$ for optimization. In other words, $\mathcal{M}$ should ``traverse'' several basins of attractions of $f$. Loosely speaking, in contrast to a linear sub-manifold (e.g., our previous work~\cite{RaponiWBBD20}), the non-linear one would solve the issue that the correlation among search variables is non-linear (e.g., on multi-modal functions), where it is challenging to identify a linear sub-manifold that passes through several local optima simultaneously. KPCA tackles this issue by first casting the search points to a high-dimensional Hilbert space $\H$ (typically infinite-dimensional), where we learn a linear sub-manifold thereof.
We consider a positive definite function $k\colon \domain \times \domain \rightarrow \R$, which induces a reproducing kernel Hilbert space (RKHS) \H constructed as the completion of $\Span{k(\BF{x}, \cdot)\colon \BF{x}\in\domain}$. 
The function $\phi(\BF{x})\coloneqq k(\BF{x}, \cdot)$ maps a point $\BF{x}$ from the search space to \H, which we will refer as \emph{the feature map}. An inner product on \H is defined with $k$, i.e., $\forall\BF{x},\BF{x}'\in\domain, \  \inner{\phi(\BF{x})}{\phi(\BF{x}')} = k(\BF{x}, \BF{x}')$, known as the \emph{kernel trick}.\\
\paragraph{The KPCA-BO algorithm.} 
Fig.~\ref{fig:kpcabo} provides an overview of the proposed KPCA-BO algorithm. 
We also present the pseudo-code of KPCA-BO in Alg.~\ref{alg:kpcabo}. Key differences to our previous work that employs the linear PCA method~\cite{RaponiWBBD20} are highlighted. 
\begin{figure}[h]
    \centering
    \includegraphics[trim=0 30 0 30,clip,width=0.95\textwidth]{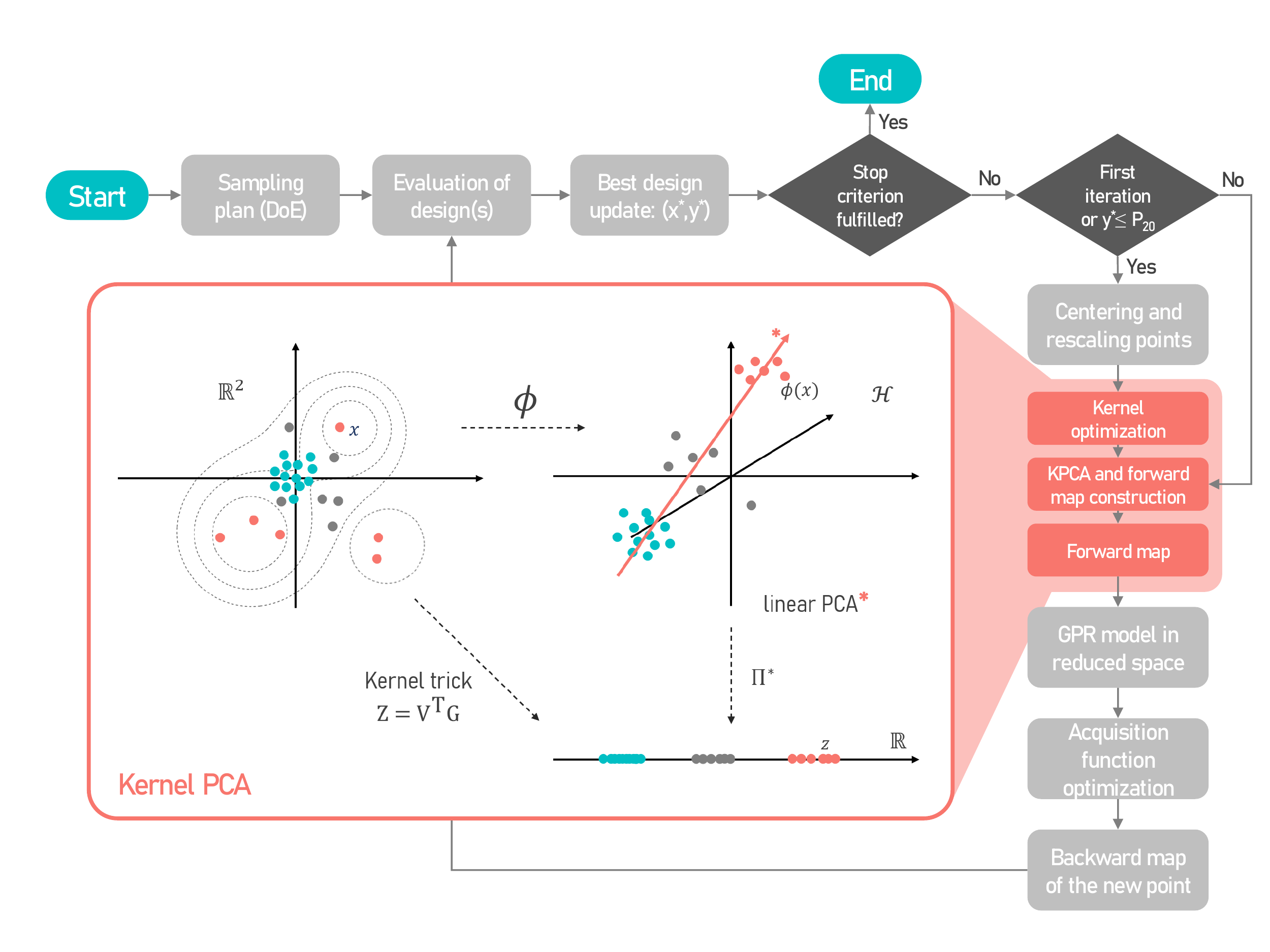}
    \caption{Flowchart of the KPCA-BO optimization algorithm with detailed graphical representation of the KPCA subroutine.}
    \label{fig:kpcabo}
\end{figure}
Various building blocks of the algorithm will be described in the following paragraphs. Notably, the sub-routine $\textsc{kpca}$ indicates performing the standard kernel PCA algorithm, which returns a set of selected principal components, whereas, $\textsc{gpr}$ represents the training of a Gaussian process regression model. In the following discussion, we shall denote by $\X=\{\BF{x}_i\}_{i=1}^n$ and $\Y=\{f(\BF{x}_i)\}_{i=1}^n$ the set of the evaluated points and their function values, respectively.\\
\paragraph{Rescaling data points.} 
As an unsupervised learning method, KPCA disregards the distribution of the function values if applied directly, which contradicts our aim of capturing the objective function in the dimensionality reduction. To mitigate this issue, we apply the weighting scheme proposed in~\cite{RaponiWBBD20}, which scales the data points in \domain with respect to their objective values. In detail, we compute the rank-based weights for all points: $w_i$ is proportional to $\ln{n} - \ln{R_i}, i=1,\ldots, n$, where $R_1, R_2,\ldots, R_n$ are the rankings of points with respect to \Y in increasing order (minimization is assumed). Then we rescale each point with its weight, i.e., $\BF{x}_i =  w_i(\BF{x}_i - n^{-1}\sum_{k=1}^n \BF{x}_k), i=1,\ldots, n$. It is necessary to show that the feature map~$\phi$ respects the rescaling operation performed in \domain. For any stationary and monotonic kernel (i.e., $k(\BF{x}, \BF{y}) = k(D_\domain(\BF{x}, \BF{y}))$ ($D_\domain$ is a metric in \domain) and $k$ decreases whenever $D_\domain(\BF{x}, \BF{y})$ increases), for all $\BF{x},\BF{y},\BF{c}\in\X$ it holds that $D_\domain(\BF{x},\BF{c}) \leq D_\domain(\BF{y},\BF{c})$ implies that $D_\H(\phi(\BF{x}),\phi(\BF{c})) \leq D_\H(\phi(\BF{y}),\phi(\BF{c}))$. Consequently, the point pushed away from the center of the data in \X will still have a large distance to the center of the data in \H after the feature map. 
The rescaling (or alternatives that incorporate the objective values into the distribution of data points in the domain) is an essential step in applying PCA to BO. 
\begin{algorithm}[!t]
	\caption{KPCA-assisted Bayesian Optimization. Highlighted are those lines in which KPCA-BO differs from the linear PCA method~\cite{RaponiWBBD20}}\label{alg:pca-bo}
	\begin{algorithmic}[1]
		\Procedure{kpca-bo}{$f,\domain$} 
		\Comment{$f$: objective function, $\domain \subseteq \R^d$: search space}
		\State{Create $\X = \{\mathbf{x}_1,\mathbf{x}_2,\ldots,\mathbf{x}_{n_0}\} \subset \domain$ with Latin hypercube sampling}
		\State{$\Y \leftarrow \{f(\mathbf{x}_1),\ldots,f(\mathbf{x}_{n_0})\}, \; n \leftarrow n_0$}
		\While{the stop criteria are not fulfilled}
            \If{$n = n_0$ or $y^*\leq$ 20\%-percentile of $\Y$}
                \State{$R_1, R_2,\ldots, R_n$ are the rankings of points in \X w.r.t. \Y (increasing order)}
        		\State{$\BF{x}_i' \gets \BF{x}_i - n^{-1}\sum_{k=1}^n \BF{x}_k, i=1,\ldots, n$} \Comment{centering}
        		\State{$\BF{x}_i' \gets w_i\BF{x}_i', w_i \propto \ln n- \ln R_i, i=1,\ldots, n$} \Comment{rescaling}
                \State{\HiLii$\gamma^*\leftarrow \textsc{optimize-rbf-kernel}(\{\BF{x}_i'\}_{i=1}^n)$} \Comment{Eq.~\eqref{eq:optimize-gamma}}
            \EndIf
            \State{\HiLi$v_1,\ldots, v_r \leftarrow \textsc{kpca}(\{\BF{x}_i'\}_{i=1}^n, \gamma^*)$}  \Comment{$r < d \ll n$}
            \State{\HiLi construct the forward map $\mathcal{F}$ from $\Span{v_1,\ldots, v_r}$. }\Comment{Eq.~\eqref{eq:forward-map}}
            \State{\HiLi{$\BF{z}_i \leftarrow \mathcal{F}(\BF{x}_i), i=1,\ldots, n$}} \Comment{map the data to $T\coloneqq\Span{v_1,\ldots, v_r}$}
		    \State{$\hat{f}, \hat{s}^2 \leftarrow \textsc{gpr}(\{\BF{z}_i\}_{i=1}^n,\Y)$} \Comment{Gaussian process regression}
	 		\State{$\BF{z}^* \leftarrow \argmax_{\BF{z}\in T}\operatorname{EI}(\BF{z};\hat{f}, \hat{s}^2)$} 
	 		\State{\HiLi$\BF{x}^* \gets \mathcal{B}(\BF{z}^*)$} \Comment{the backward map; Eq.\eqref{eq:backward-map}}
	 		\State{$y^* \gets f(\BF{x}^*)$}
	 		\State{$\X \gets \X \cup \{\BF{x}^*\}, \;\Y \gets \Y \cup \{y^*\}, \; n\leftarrow n + 1$}
	 	\EndWhile
	 	\EndProcedure
	\end{algorithmic}
	\label{alg:kpcabo}
\end{algorithm}
On the one hand, DoE aims to span the search domain as evenly as possible and thereby the initial random sample from it has almost the same variability in all directions, which provides no information for PCA to learn. On the other hand, new candidates are obtained by the global optimization of the acquisition function in each iteration, which is likely to produce multiple clusters and/or isolated points that are not meaningful to the PCA procedure. This is in contrast to the direct application of PCA to evolutionary algorithms~\cite{kapsoulis_use_2016}, where we apply PCA to the current population. Since the population is usually generated from a unimodal mutation distribution, it is well-suited for applying the PCA procedure.\\
\paragraph{Dimensionality reduction in Hilbert spaces.}  
After the rescaling operation in \X, we map the points to the feature space \H: $\phi(\X)=\{\phi(\BF{x}_i)\}_i,1=1,\ldots, n$. After centering the feature points in \H, i.e., $
\tilde{\phi}(\BF{x}_i) = \phi(\BF{x}_i) - n^{-1}\sum_{i=1}^n \phi(\BF{x}_i)$, we express the sample covariance\footnote{The outer product is a linear operator defined as $\forall h\in\H, [\phi(\BF{x})\phi(\BF{x})^\top](h)\colon h\mapsto \langle \phi(\BF{x}), h \rangle_{\H}\phi(\BF{x})$. Hence, the sample covariance is also a linear operator $C\colon \H \rightarrow \H$.} of the feature points: $C = n^{-1}\sum_{i=1}^n \tilde{\phi}(\BF{x}_i)\tilde{\phi}(\BF{x}_i)^\top$. KPCA essentially computes the eigenvalues and eigenfunctions of $C$, namely $\forall i\in [1..n], Cv_i =\lambda_iv_i, v_i \in \H, ||v_i||_\H = 1$, and $\langle v_i, v_j\rangle_\H = 0$, if $i\neq j$. Note that (1) $C$ is positive semi-definite; (2) since $\rank(C) \leq \sum_{i=1}^{n}\rank(\tilde{\phi}(\BF{x}_i)\tilde{\phi}(\BF{x}_i)^\top) = n$, there are maximally $n$ nonzero eigenvalues and eigenfunctions; (3) the eigenfunction takes the following form $v_i = \sum_{j=1}^n a_j^{(i)} \tilde{\phi}(\BF{x}_j), a_j^{(i)}\in\R$ and thereby all eigenfunctions can be represented by a matrix $\BF{V} = (a_j^{(i)})_{ij}$. Assume the eigenvalues are ordered in the decreasing manner (i.e., $\lambda_1 \geq \lambda_2 \geq \cdots \geq \lambda_n \geq 0$. Eigenfunctions are sorted accordingly). It is not hard to show that the variance of $\phi(\X)$ along $v_i$ is exactly $\lambda_i$:
$n^{-1}\sum_{k=1}^n\langle\tilde{\phi}(\BF{x}_k),v_i\rangle_\H^2 = \langle v_i, n^{-1}\sum_{k=1}^n[\tilde{\phi}(\BF{x}_k)\tilde{\phi}(\BF{x}_k)^\top]v_i\rangle_\H = \lambda_i.$
Therefore, the eigenvalues can be used to select a linear subspace of \H which keeps the majority of the variability of $\phi(\X)$. Specifically, we choose a subspace $T\coloneqq\Span{v_1,\ldots,v_r} \subset \H$ as the reduced search space of BO, where $r$ is chosen as the smallest integer such that the first-$r$ eigenvalues explain at least $\eta$~percent of the total variability (we use $\eta = 90\%$ in our experiments). For a point $\BF{x}\in\domain$, we can formulate a \emph{forward map} that projects $\phi(\BF{x})$ onto the reduced space $T$: $\mathcal{F}\colon \BF{x} \mapsto \sum_{i=1}^r\langle\tilde{\phi}(\BF{x}), v_i\rangle_\H v_i$. Let $g_i(\BF{x}) = \langle\tilde{\phi}(\BF{x}), \tilde{\phi}(\BF{x}_i)\rangle_\H = k(\BF{x}, \BF{x}_i) - n^{-1}\sum_{j=1}^n k(\BF{x},\BF{x}_j)- n^{-1}\sum_{j=1}^n k(\BF{x}_i,\BF{x}_j) + n^{-2}\sum_{i=1}^n\sum_{j=1}^nk(\BF{x}_i,\BF{x}_j)$, the forward map can be re-expressed as 
\begin{equation} \label{eq:forward-map}
    \mathcal{F}\colon \BF{x} \mapsto \BF{V}\BF{g}(\BF{x}), \quad \BF{g}(\BF{x}) = (g_1(\BF{x}),\ldots, g_n(\BF{x}))^\top.
\end{equation}
Computationally, the eigenfunction representation $\BF{V}$ can be calculated via eigendecomposition of the Gram matrix $(\BF{G}_{ij}=\langle\tilde{\phi}(\BF{x}_i),\tilde{\phi}(\BF{x}_j)\rangle_\H$, i.e., $\BF{G}=\BF{V}^\top\BF{D}\BF{V}$, $\BF{D}$ is a $n\times n$ diagonal matrix with the eigenvalues of $C$ on its nonzero entries.

\paragraph{Learning the forward map.} 
We use the \emph{radial basis function} (RBF, a.k.a.~Gaussian kernel) for KPCA in this paper.
The RBF kernel $k(\BF{x}, \BF{x}')=\exp(-\gamma\norm{\BF{x} - \BF{x}'}_2^2)$, contains a single length-scale hyperparameter $\gamma\in \R_{>0}$. To determine this length-scale, we minimize the number of eigenvalues/functions chosen to keep at least $\eta$ percent of the variance, which effectively distributes more information of $\phi(\X)$ on the first few eigenfunctions and hence allows for constructing a lower-dimensional space $T$. Also, we reward $\gamma$ values which choose the same number of eigenfunctions and also yield a higher ratio of explained variance. In all, the cost function for tuning $\gamma$ is: 
\begin{equation} \label{eq:optimize-gamma}
    \gamma^* = \argmin_{\gamma \in (0,\infty)} r - \frac{\sum_{i=1}^r \lambda_i}{\sum_{i=1}^n \lambda_i}, \quad r=\inf\left\{k\in[1..n]\colon\sum_{i=1}^k \lambda_i \geq \eta \sum_{i=1}^{n} \lambda_i\right\}.
\end{equation}
This equation is solved numerically, using a quasi-Newton method (the L-BFGS-B algorithm~\cite{ByrdLNZ95}) with $\gamma\in [10^{-4}, 2]$ and maximally $200d$ iterations. It is worth noting that we do not consider anisotropic kernels (e.g., individual length-scales for each search variable) since such a kernel increases the number of hyperparameters to learn.

Also, note that the choice of the kernel can affect the smoothness of the manifold in the feature space \H, e.g., the Mat\'ern 5/2 kernel induces a $C^{2}$ atlas for the manifold $\phi(S)$. We argue that the smoothness of $\phi(S)$ is less important to the dimensionality reduction task, comparing to the convexity and connectedness thereof. 
In this work, we do not aim to investigate the impact of the kernel on the topological properties of $\phi(S)$. Therefore, we use the RBF kernel for the construction of the forward map for its simplicity.

\paragraph{Learning the backward map.}
When performing the Bayesian optimization in the reduced space $T$, we need to determine a ``pre-image'' of a candidate point $\BF{z}\in T$ for the function evaluation. To implement such a backward map $\mathcal{B}\colon T \rightarrow \domain$, we base our construction on the approach proposed in~\cite{abs-2001-01958}, in which the pre-image of a point $\BF{z}\in T$ is a conical combination of some points in \domain: $\sum_{i=1}^{d} w_i \BF{p}_i,w_i\in \R_{>0}$. In this paper, the points $\{\BF{p}_i\}_{i=1}^d$ are taken as a random subset of the data points $\{\BF{x}_i\}_{i=1}^{n}$. The conical weights are determined by minimizing the distance between $\BF{z}$ and the image of the conical combination under the forward map:
\begin{align} 
    w_1^*, \ldots, w_d^* &= \argmin_{\{w_i\}_{i=1}^d \subset \R_{>0}^d} \norm{\BF{z} - \mathcal{F}\left(\sum_{i=1}^{d} w_i \BF{p}_i\right)}_2^2 + Q\left(\sum_{i=1}^{d} w_i \BF{p}_i\right), \label{eq:optimize-weights}\\
    Q(\BF{x})&=\exp\left(\sum_{i=1}^d\max(0, l_i - x_i) + \max(0, x_i - u_i)\right). \nonumber
\end{align}
where the function $Q$ penalizes the case that the pre-image is out of \domain. As with Eq.~\eqref{eq:optimize-gamma}, the weights are optimized with the L-BFGS-B algorithm (starting from zero with $200d$ maximal iterations). Taking the optimal weights, we proceed to define the \emph{backward map}: $\forall \BF{z} \in T$, 
\begin{equation} \label{eq:backward-map}
    \mathcal{B}\colon \BF{z} \mapsto \operatorname{CLIP}\left(\sum_{i=1}^d w_i^* \BF{p}_i\right),
\end{equation}
where the function $\operatorname{CLIP}(\BF{x})$ cuts off each component of $\BF{x}$ at the lower and upper bounds of \domain, which ensures the pre-image is always feasible.\\
\emph{Remark.} 
The event that $\{\BF{p}_i\}_{i=1}^d$ contains a co-linear relation is of measure zero and the conical form can procedure pre-images outside \domain, allowing for a complete coverage thereof.
There exist multiple solutions to Eq.~\eqref{eq:optimize-weights} (and hence multiple pre-images) since the forward map $\mathcal{F}$ contains an orthogonal projection step, which is not injective. Those multiple pre-images can be obtained by randomly restarting the quasi-Newton method used to solve Eq.~\eqref{eq:optimize-weights}. However, since those pre-images do not distinguish from each other for our purpose, we simply take a random one in this work.

\paragraph{Bayesian optimization in the reduced space.} Given the forward and backward maps, we are ready to perform the optimization task in the space $T$. Essentially, we first map the data set $\X\subset S$ to $T$ using the forward map (Eq.~\eqref{eq:forward-map}): $\mathcal{F}(\X) = \{\mathcal{F}(\BF{x}_i)\}_{i=1}^n$, which implicitly defines the counterpart $f'\coloneqq f \circ \mathcal{B}$ of the objective function in $T$. Afterwards, we train a Gaussian process model with the data set $(\mathcal{F}(\X), \Y)$ to model $f'$, i.e., $\forall \BF{z}\in T, f'(\BF{z})\mid \mathcal{F}(\X), \Y, \BF{z} \sim \mathcal{N}(\hat{f}(\BF{z}), \hat{s}^2(\BF{z}))$. 
The search domain in the reduced space $T$ is determined as follows. Since the RBF kernel monotonically decreases w.r.t. the distance between its two input points, we can bound the set $\phi(\X)$ by first identifying the point $\BF{x}_{\text{max}}$ with the largest distance to the center of data points $\BF{c}$ and secondly computing the distance $r$ between $\phi(\BF{x}_{\text{max}})$ and $\phi(\BF{c})$ in the feature space \H. Since $S$ is a hyperbox in $\mathbb{R}^d$, we simply take an arbitrary vertex of the hyperbox for $\BF{x}_{\text{max}}$. Note that, as the orthogonal projection (from $\H$ to  $T$) does not increase the distance, the open ball $B\coloneqq \{\BF{z}\in T\colon \norm{\BF{z}}_2 < r\}$ always covers $\mathcal{F}(\X)$. For the sake of optimization in $T$, we take the smallest hyperbox covering $B$ as the search domain in $T$. 

After the GPR model is created on the data set $(\mathcal{F}(\X), \Y)$, we maximize the expected improvement function $\operatorname{EI}(\BF{z};\hat{f},\hat{s}) = \hat{s}(\BF{z})u\operatorname{CDF}(u) + \hat{s}(\BF{z})\operatorname{PDF}(u), u= (\min\Y - \hat{f}(\BF{z}))/\hat{s}(\BF{x})$ to pick a new candidate point $\BF{z}^*$, where $\operatorname{CDF}$ and $\operatorname{PDF}$ stand for the cumulative distribution and probability distribution functions of a standard normal random variable, respectively. 
Due to our construction of the search domain in $T$, it is possible that the global optimum $\BF{z}^*$ of EI is associated with an infeasible pre-image in \domain. To mitigate this issue, we propose a multi-restart optimization strategy for maximizing EI (with different starting points in each restart), in which we only take the best outcome (w.r.t. its EI value) whose pre-image belongs to \domain. In our experiments, we used $10$ random restarts of the optimization. 
Also, it is unnecessary to optimize kernel's hyperparameter $\gamma$ in each iteration of BO since the new point proposed by EI would not make a significant impact on learning the feature map, if its quality is poor relative to the observed ones in \Y (and hence assigned with a small weight). Therefore, it suffices to only re-optimize $\gamma$ whenever we find a new point whose function value is at least as good as the $20\%$ percentile of \Y. Also, the $20\%$ threshold is manually chosen to balance the convergence and computation time of KPCA-BO, after experimenting several different values on BBOB test problems.

\section{Experiments}\label{sec:experiement}
\paragraph{Experimental Setup.} 
We evaluate the performance of KPCA-BO on ten multi-modal functions from the BBOB problem set~\cite{hansen2020coco} (F15 - F24), which should be sufficiently representative of the objective functions handled in real-world applications.
We compare the experimental result of KPCA-BO to standard BO and the PCA-BO in our previous work~\cite{RaponiWBBD20} on three problem dimensions $d\in\{20, 40, 60\}$ with the evaluation budget in $\{100, 200, 300\}$, respectively. 
We choose a relatively large DoE size of $3d$, to ensure enough information for learning the first sub-manifold.
On each function, we consider five problem instances (instance ID from $0$ to $4$) and conduct 10 independent runs of each algorithm.
We select the Matérn $5/2$ kernel for the GPR model. The L-BFGS-B algorithm~\cite{ByrdLNZ95} is employed to maximize the likelihood of GPR as well as the EI acquisition function at each iteration.
We add to our comparison results for CMA-ES~\cite{hansen2001self_adaptation_es}, obtained by executing the \texttt{pycma} package (\url{https://github.com/CMA-ES/pycma}) with $16$ independent runs on each problem. The implementation of BO, PCA-BO, and KPCA-BO can be accessed at \url{https://github.com/wangronin/Bayesian-Optimization/tree/KPCA-BO}. \\
\begin{figure}[!ht]
\centering
\includegraphics[width=1\textwidth, trim=25mm 05mm 35mm 00mm, clip]{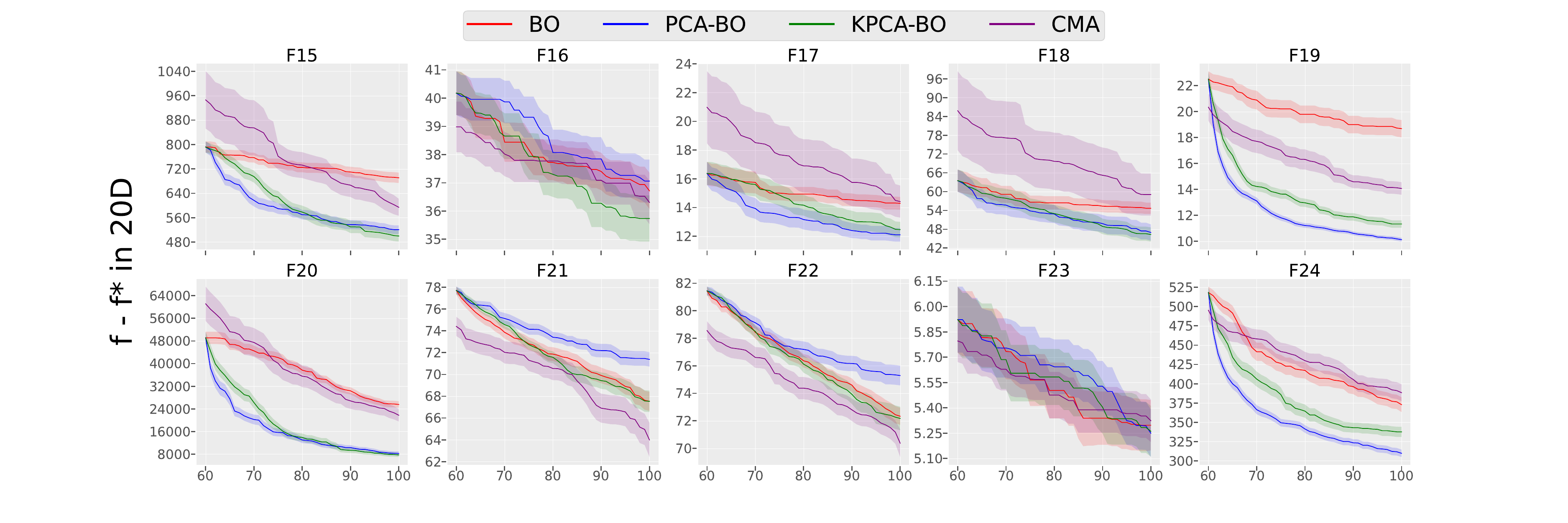}
\includegraphics[width=1\textwidth, trim=25mm 05mm 35mm 10mm, clip]{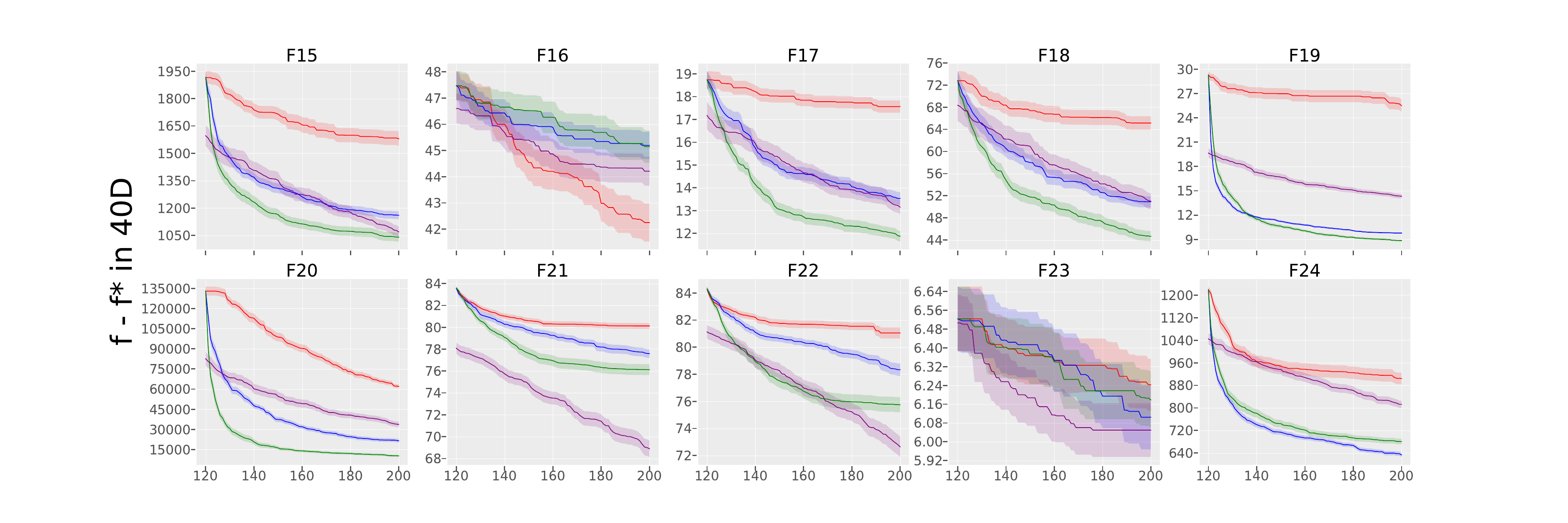}
\includegraphics[width=1\textwidth, trim=25mm 00mm 35mm 10mm, clip]{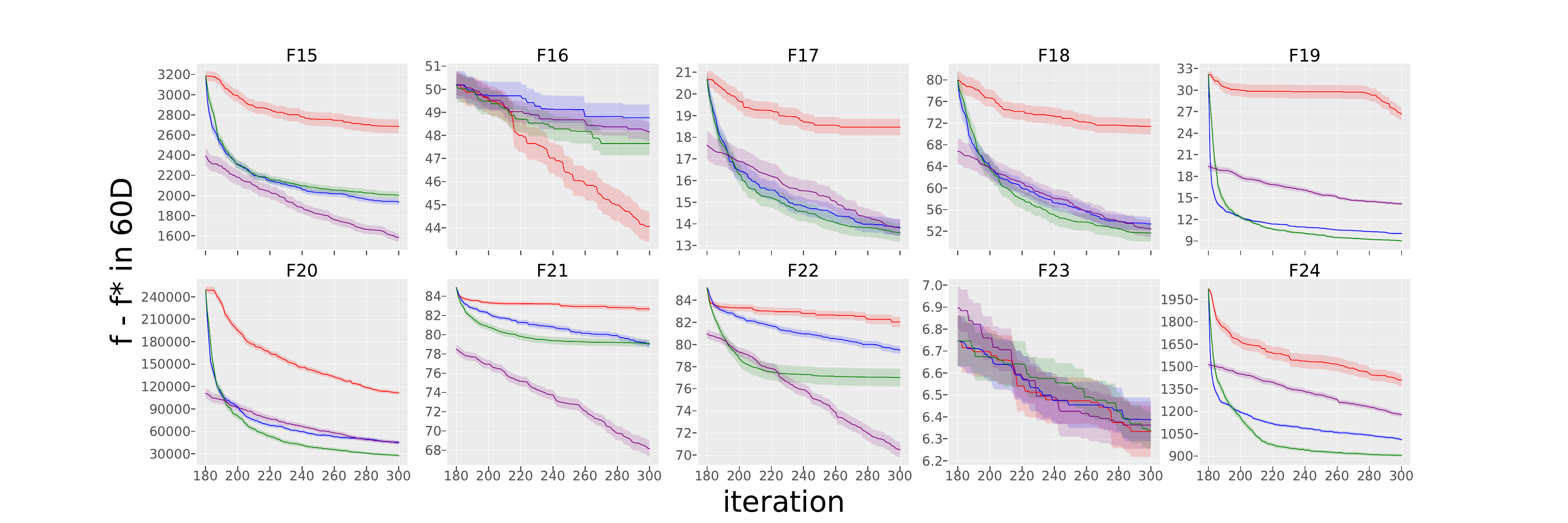}
\caption{The best-so-far target gap ($f_{\text{best}} - f^*$) against the iteration for CMA-ES (purple), BO (red), PCA-BO (blue), and KPCA-BO (green) is averaged over $50$ independent runs on each test problem. We compare the algorithms in three dimensions, $20$D (top), $40$D (middle), and $50$D (bottom). The shaded area indicates the standard error of the mean target gap. CMA-ES data is obtained from running the \texttt{pycma} package with the same evaluation budgets as BO.
}
\label{fig:convergence_plots}
\end{figure}

\paragraph{Results.} All our data sets are available for interactive analysis and visualization in the IOHanalyzer~\cite{IOHanalyzer} repository, under the \emph{bbob-largescale} data sets of the \emph{IOH} repository. In Fig.~\ref{fig:convergence_plots}, we compare the convergence behavior of all four algorithms, where we show the evolution of the best-so-far target gap ($f_{\text{best}} - f^*$) with respect to the iteration for each function-dimension pair. In all dimensions, it is clear that both KPCA-BO and PCA-BO outperform BO substantially across functions F17 - F20, while both KPCA-BO and PCA-BO exhibit about the same convergence with BO on F23, and are surpassed by BO significantly on F16.
The poor performance of PCA-BO and KPCA-BO on the Weierstrass function (F16) can be attributed to the nature of its landscape, which is highly rugged and moderately periodic with multiple global optima. Therefore, a basin of attraction is not clearly defined, which confuses both PCA variants.
On functions F17, F19, and F24, we observe that KPCA-BO is outperformed by PCA-BO in $20$D, while as the dimensionality increases, KPCA-BO starts to surpass the convergence rate of PCA-BO.
For functions F21 and F22, KPCA-BO's performance is indistinguishable from PCA-BO in $20$D, and in higher dimensions, the advantage of KPCA-BO becomes prominent. 
For the remaining function-dimension pairs, KPCA-BO shows a comparable performance to PCA-BO. Compared to CMA-ES, both KPCA-BO and PCA-BO either outperform CMA-ES or show roughly the same convergence except on F21 and F22 in $20$D. 
In higher dimensions, although KPCA-BO still exhibits a steeper initial convergence rate (before about 200 function evaluations in $60$D), CMA-ES finds a significantly better solution after the first $3d$ evaluations (the DoE phase of the BO variants), leading to better overall performance.

Also, we observe that KPCA-BO shows better relative performance when the dimensionality increases (e.g., on F18 and F22 across three dimensions), implying that the kernelized version is better suited for solving higher-dimensional problems. Interestingly, KPCA-BO shows an early faster convergence on F21 and F22 compared to PCA-BO and is gradually overtaken by PCA-BO, implying that KPCA-BO stagnates earlier than PCA-BO. We conjecture that the kernel function of KPCA-BO (and consequently the sub-manifold ) stabilizes much faster than the linear subspace employed in PCA-BO, which might attribute to such a stagnation behavior. Therefore, KPCA-BO is more favorable than PCA-BO in higher dimensions (i.e., $d \geq 60$), while in lower dimensions ($d \approx 20$), it is competitive to PCA-BO when the budget is small for most test cases (it only loses a little to PCA-BO on F24).

In Fig.~\ref{fig:CPU-time}, we depict the CPU time (in seconds) taken to train the GPR model as well as maximize EI in $60$D. As expected, the majority of the CPU time is taken by the training of the GPR model. PCA-BO and KPCA-BO achieve substantially smaller CPU time of GPR training than the vanilla BO with exception on F19, F20, and F24, where KPCA-BO actually takes more time. In all cases, the CPU time for maximizing EI is smaller for KPCA-BO and PCA-BO than for BO.
\begin{figure}[t]
\centering
\includegraphics[width=0.95\textwidth, trim=0mm 0mm 0mm 00mm, clip]{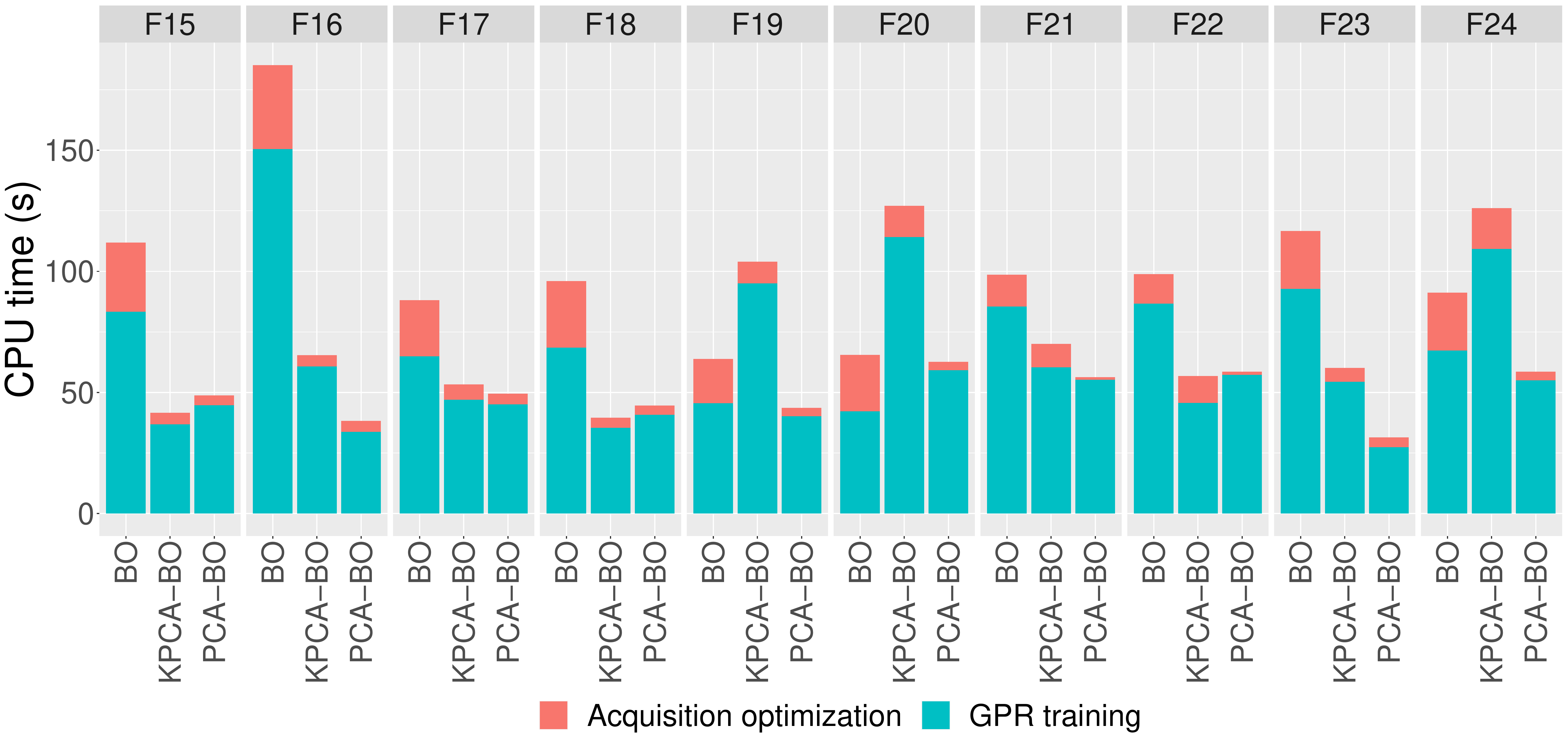} \\
\caption{Mean CPU time taken by training the GPR model (dark cyan) and maximizing the $\operatorname{EI}$ acquisition function (red) in $60$D for BO, PCA-BO, and KPCA-BO, respectively. In general, training the GPR model takes the majority of CPU time and both PCA-BO and KPCA-BO manages to reduce it significantly.}
\label{fig:CPU-time}
\end{figure}
\section{Conclusions}\label{sec:conclusion}
In this paper, we proposed a novel KPCA-assisted Bayesian optimization algorithm, the  KPCA-BO. Our algorithm enables BO in high-dimensional numerical optimization tasks by learning a nonlinear sub-manifold of the original search space from the evaluated data points and performing BO directly in this sub-manifold. Specifically, to capture the information about the objective function when performing the kernel PCA procedure, we rescale the data points in the original space using a weighting scheme based on the corresponding objective values of the data point. With the help of the KPCA procedure, the training of the Gaussian process regression model and the acquisition optimization -- the most costly steps in BO -- are performed in the lower-dimensional space. We also implement a backward map to convert the candidate point found in the lower-dimensional manifold to the original space.

We empirically evaluated KPCA-BO on the ten multimodal functions (F15-F24) from the BBOB benchmark suite. We also compare the performance of KPCA-BO with the vanilla BO, the PCA-BO algorithm from our previous work, and CMA-ES, a state-of-the-art evolutionary numerical optimizer. The results show that KPCA-BO performs better than PCA-BO in capturing the contour lines of the objective functions when the variables are not linearly correlated. The higher the dimensionality, the more significant this better capture becomes for the optimization of functions F20, F21, F22, and F24. 
Also, the mean CPU time measured in the experiments shows that after reducing the dimensionality, the CPU time needed to train the GPR model and maximize the acquisition is greatly reduced in most cases.

The learning of the lower-dimensional manifold is the crux of the proposed KPCA-BO algorithm. However, we observe that this manifold stabilizes too quickly for some functions, leading to unfavorable stagnation behavior. In further work, we plan to investigate the cause of this premature convergence and hope to identify mitigation methods. Since the manifold is learned from the data points evaluated so far, a viable approach might be to use a random subset of data points to learn the manifold, rather than taking the entire data set.

Another future direction is to improve the backward map proposed in this paper. Since an orthogonal projection is involved when mapping the data to the manifold, the point on the manifold has infinitely many pre-images. 
In our current approach, we do not favor one direction or another, whereas it might be preferable to bias the map using the information of the previously sampled points. For example, the term to minimize (to exploit) or maximize (to explore) the distance between the candidate pre-image point and the lower-dimensional manifold can be added to the cost function used in the backward map. These two approaches can be combined or switched in the process of optimization.

\vspace{1ex}
\textbf{Acknowledgments. } 
Our work is supported by the Paris Ile-de-France region (via the DIM RFSI project AlgoSelect), 
by the CNRS INS2I institute (via the RandSearch project), 
by the PRIME programme of the German Academic Exchange Service (DAAD) with funds from the German Federal Ministry of Education and Research (BMBF), and by RFBR and CNRS, project number 20-51-15009.
}

\newcommand{\etalchar}[1]{$^{#1}$}

\end{document}